\documentclass{article}

\usepackage{microtype}
\usepackage{graphicx}
\usepackage{subcaption}
\usepackage{booktabs}
\usepackage{hyperref}

\usepackage[accepted]{icml2026}
\usepackage{amsmath}
\usepackage{amssymb}
\usepackage{mathtools}
\usepackage{amsthm}
\usepackage[capitalize,noabbrev]{cleveref}
\usepackage[textsize=tiny]{todonotes}
\usepackage{placeins}
\usepackage{makecell}

\usepackage{tikz}        
\usetikzlibrary{positioning, fit, shapes.geometric, arrows.meta, calc, backgrounds, decorations.pathreplacing}
\usepackage{amsmath, amssymb}
\usepackage{enumitem}

\icmltitlerunning{\modelname: Efficient In-Context Learning for Tabular Prediction}

\usepackage{xspace}
\newcommand{\modelname}{TabDPT-Turbo\xspace}

\newcommand{\pgraph}[1]{\textbf{#1}\,\,}

\begin{document}

\twocolumn[
  \icmltitle{\modelname: Efficient In-Context Learning for Tabular Prediction}

  \icmlsetsymbol{equal}{*}

  \begin{icmlauthorlist}
    \icmlauthor{Rasa Hosseinzadeh}{l6}
    \icmlauthor{Alex Labach}{l6}
    \icmlauthor{Zexin Xue}{l6}
    \icmlauthor{Shuyi Han}{l6}
    \icmlauthor{Valentin Thomas}{cohere}
    \icmlauthor{Anthony L. Caterini}{l6}
  \end{icmlauthorlist}

  \icmlaffiliation{l6}{Layer 6 AI, Toronto, Canada}
  \icmlaffiliation{cohere}{Cohere, Toronto, Canada, work done while at Layer 6 AI}

  \icmlcorrespondingauthor{}{\{rasa,alex,anthony\}@layer6.ai}

  \icmlkeywords{Tabular Learning, In-Context Learning, Foundation Models}

  \vskip 0.3in
]

\printAffiliationsAndNotice{}

\begin{abstract}

Tabular foundation models, driven by in-context learning, have rapidly grown in quality and popularity.
However, recent approaches with either cell-based architectures or retrieval have sacrificed efficiency for raw performance, restricting their utility in situations where compute is limited or inference speed is crucial.
We adopt an alternate approach, sticking with row-based attention while incorporating long context pre-training to eliminate the need for retrieval. 
By combining this with architectural improvements and SSL pre-training on a newly-sourced, larger corpus of real data results, we present \modelname, a model that provides comparable default performance to TabDPT v1.1 on TabArena-Lite, CC18, and CTR23, at orders of magnitude faster. In our experiments, \modelname is the fastest model overall among leading foundation models. We have released the new model as TabDPT v1.2 at \url{https://github.com/layer6ai-labs/TabDPT-inference}.

\end{abstract}

\section{Introduction}

\looseness=-1
Tabular data remains the primary modality driving predictive AI in industry~\citep{breugel2024tabular}.
Tabular Foundation Models (TFMs), backed by in-context learning (ICL) have emerged in response, offering high-quality predictions without the need for extensive model training or hyperparameter tuning~\citep{hollmann2023tabpfn,qu2025tabicl}.
However, many recent TFMs rely on expensive operations such as retrieval~\citep{ma2025tabdpt, zhang2025limix} or cell-based attention~\citep{hollmann2025accurate} to push performance at the cost of efficiency.
This reduces their scope, limiting effectiveness in resource-constrained environments or low-latency settings. %

This work instead revisits row-based TFMs~\citep{hollmann2023tabpfn}.
Efficiency and low resource deployment are typically easier to attain with row-based rather than cell-based architectures, particularly as context length increases.
Besides the immediate benefits, enabling a more performant architecture with efficiency as a core design principle also provides the flexibility to perform inference-time adjustments such as fine-tuning and larger-scale explainability.

\looseness=-1
Considering these benefits, we directly extend TabDPT~\citep{ma2025tabdpt} as it is the highest-performing row-based architecture on the standard TabArena benchmark~\citep{erickson2025tabarena} and is fully open-sourced.
We pre-train on real data with the same Self-Supervised Learning (SSL) procedure, albeit with a corpus sourced from OpenML~\citep{vanschoren2014openml} which is an order of magnitude larger than the previous pre-training set used in TabDPT.
We also incorporate numerous architecture changes to improve model performance, including optimizations specifically developed for long-context training which enables us to avoid retrieval.
While retrieval has been shown to improve TFM performance by providing a tailored local context to each test example~\citep{retrieve2024thomas,ma2024context}, it also greatly increases inference compute and memory requirements.
Furthermore, the relevance of retrieval becomes less clear in the presence of long contexts; it has been shown in concurrent work~\citep{shaheen2026generalization} that ICL-based TFMs learn nearest-neighbour-like behaviour internally.
Our resulting model, \modelname, is an accelerated version of TabDPT with matching or better performance compared to previous versions.
We summarize our contributions below:
\begin{itemize}[noitemsep,nolistsep]
  \item We present \modelname: a retrieval free, long context, row-based TFM aimed at efficient inference.
  \item We provide a new pre-training data corpus based on filtered and deduplicated OpenML tables, preserving the TabDPT-style SSL objective while scaling the amount and diversity of training data.
  \item We introduce numerous architecture and loss changes: attention temperature scaling, per-layer target routing through transformer value projections, learned thinking rows, regression-as-classification with Continuous Ranked Probability Score (CRPS), and an auxiliary context prediction loss.
  \item Results on TabArena-Lite~\cite{erickson2025tabarena}, CC18~\cite{bischl2021cc18}, and CTR23~\cite{fischer2023ctr} show better predictive performance than TabDPT v1.1 at \textbf{orders of magnitude faster} inference, resulting in the fastest model among leading TFMs.
\end{itemize}

\section{Related Work}

\looseness=-1
\pgraph{Tabular Foundation Models} TFMs are an active area of research, becoming dominant in tabular predictive benchmarks~\cite{erickson2025tabarena}, with TabPFN variants~\cite{hollmann2023tabpfn,hollmann2025accurate}, TabICL and TabICLv2 
~\cite{qu2025tabicl,qu2026tabiclv2}, TabDPT~\cite{ma2025tabdpt}, and many other strong models being introduced recently~\cite{zhang2025mitra,zhang2025limix,bouadi25orionmsp}. But for a given compute budget, traditional or supervised tabular models can still provide competitive trade-offs, with Random Forest~\cite{breiman2001random}, EBM~\cite{lou2013accurate}, CatBoost~\cite{prokhorenkova2018catboost}, and TabM~\cite{gorishniy2024tabm} all appearing on TabArena Pareto curves at the time of writing. A number of major design decisions affect TFM compute usage, and so we approach them with a focus on efficiency.

\looseness=-1
\pgraph{Cell- vs.\ Row-Based Architectures}
Many recent TFMs opt for cell-based attention mechanisms~\citep{grinsztajn2025tabpfn, zhang2025limix} instead of the classic row-based architecture popularized by TabPFNv1~\citep{hollmann2023tabpfn}.
Indeed, a cell-based tokenization is expressive for heterogeneous columns and gets closer to encoding column invariance, which is often considered a desirable property for TFMs, but sequence length scales with $n \times m$ for $n$ rows and $m$ columns.
We instead opt for a row-based setup as a deliberate efficiency choice: since each row is one token, long contexts become more feasible and inference remains easier to batch.
\citet{qu2026tabiclv2} use a hybrid approach, first using a lower-dimensional cell-based transformer that essentially acts as an encoder before passing through a row-based transformer; this is still slower than fully row-based models when the number of features increases.

\looseness=-1
\pgraph{Retrieval}
Several existing models, including TabDPT, use retrieval to select instances to include in the context~\cite{retrieve2024thomas, ma2025tabdpt,zhang2025limix}. This operation allows the model to utilize more training instances, but requires a separate context for each inference instance, meaning instances cannot be batched along with a shared context.
Using long contexts instead maintains approximate locality in TFMs via attention~\citep{shaheen2026generalization} and so we elect to remove the retrieval for more efficient inference.

\section{Method}

\subsection{Data}

\looseness=-1
We adopt TabDPT's general training approach as a starting point since it is an open source row-based architecture.
While TabDPT was originally trained on 112 datasets, the authors argued for the potential of scaling data to improve model performance. We follow this direction by scaling to \textbf{1,445} datasets. We also deduplicate with respect to TabArena, which was not done in previous TabDPT versions, providing additional robustness to our results.

\looseness=-1
To extend the training data, we retrieved \href{https://www.openml.org/search?type=data&sort=runs&status=active}{all active datasets on OpenML}~\cite{OpenML2025} and applied filtering and deduplication stages to derive our final training set.
First, we used the deduplication code provided in the original TabDPT training repository to exclude all datasets that were possible duplicates or derivations of datasets in CC18, CTR23, and TabArena. %
Empirically, we observed that datasets with too few columns could harm performance, and therefore filtered out datasets with fewer than 10 columns. We also removed excessively large datasets to speed up training, as we did not observe any significant change in final performance when doing so. Specifically, we removed datasets with over 200 columns or a 300 MB file size.
We also applied column-level filters as part of pre-processing, removing categorical columns with a cardinality over 100, and entirely constant columns. 

Finally, we added another duplicate detection step among groups of datasets that had equal row and column count. We reordered each candidate table to make comparisons invariant to row and column order.
Rows and columns were permuted so that the largest element appeared in the top-left corner,
after which the first row and first column were sorted.
If two reordered tables were within a fixed tolerance, we kept only one of them.
We compare our pre-training corpus with TabDPT v1.1 in~\Cref{tab:v1_vs_v2} in the appendix.

Beyond scaling for performance, extending the training data ensures that we have a diverse range of datasets (up to 5M rows), enabling us to train the model with longer context.

\pgraph{Self-Supervised Setup}
Our SSL procedure follows TabDPT v1.1~\citep{ma2025tabdpt}.
There, a table is first randomly sampled, then a task is constructed from this table. A column is selected, and if it satisfies some basic quality checks, it can be used as either a regression or classification target. Some post-processing is used to introduce more diversity, such as randomizing/merging classes or applying random functions and normalizing for regression. A random subset of the remaining columns is used as features. Finally, a random subset of instances is selected to be used as context, and another as queries; the TFM is trained to predict the targets for the queries given the query features, along with the context features and targets. In this work we use context lengths up to 32k rows.

\subsection{Architecture}

We use a row-based transformer architecture for efficiency. Each table row is
padded with zeros to 128 dimensions and then encoded with a linear layer. This
follows the TabDPT design (with 128 instead of 100) and avoids the cost of cell-level modelling, whose
sequence length scales with both rows and columns. The trade-off is that the
encoder is not intrinsically invariant to column order. However, the row-based design is
more efficient, stable during training, and robust against representation
collapse. It enables ensembling by permuting feature columns. For datasets with
more than 128 features, the model must reduce the feature dimension with
methods like PCA or column subsampling.

\looseness=-1
The transformer backbone is a pre-norm variant using a 512-dimensional embedding,
32 transformer layers, 8 attention heads, and SwiGLU blocks. We also prepend 64
learned thinking rows to the sequence as in~\citet{hollmann2025accurate}. These rows are soft tokens and are not
passed through the linear encoder; instead, they participate in attention, providing
additional latent computation. Following TabDPT's attention design, rows attend only to
the context and thinking rows, preventing query-to-query information flow.

Target conditioning is injected inside each transformer layer rather than being
simply added to the row embedding as opposed to TabDPT. Each layer has a small
target encoder that maps context targets through an MLP to embeddings. These
embeddings are concatenated to the corresponding context row representations in
the attention value stream, while queries and keys remain functions of the row
representations. This makes target information available to the model through
values while keeping context and query row representations similar to each
other.

Queries and keys are additionally normalized per head before scaled
dot-product attention. The attention branch includes a learned sigmoid gate, 
computed per token and attention head, which multiplicatively gates the
attention output before the output projection. We also scale the attention
temperature as the context length grows to mitigate dispersion~\cite{velickovic2025softmax}.

\looseness=-1
The prediction head is an MLP with one hidden layer that maps the final row
representation to a joint output vector with 16 classification logits and
 2048 regression logits. For classification, the first entries are
interpreted as class logits and normalized with a softmax over the active
classes. For regression, the remaining logits parameterize a categorical
distribution over uniform bins spanning 10 standard deviations around the mean
(calculated from context samples). The softmax over these bins gives a
predictive distribution, and the point prediction is obtained as the expected
bin centre. We depict the architecture in Figure~\ref{fig:arch-main} in the Appendix.

\subsection{Objective}

For classification tasks, we train with cross-entropy over up to 16 classes,
matching the first 16 logits of the joint prediction head. When a dataset has
fewer than 16 classes, only the active class logits are used. For datasets with
more than 16 classes during inference, prediction can be extended using the
same digit-by-digit strategy as TabDPT.

For regression, targets are standardized, and the model predicts a categorical
distribution over 2048 bins spanning the interval $[-10, 10]$ in standardized
target space (akin to~\citet{balazadeh2025causalpfn} and \citet{hollmann2025accurate}). We optimize the CRPS loss between the predicted cumulative
distribution and the target CDF induced by the observed value. This avoids
choosing an arbitrary smoothing bandwidth required by cross-entropy on the same
task. The CRPS loss is a proper scoring rule~\citep{landsgesell2026scoringbench} which encourages calibrated predictive
distributions, offering more information than TabDPT's point estimates. At
inference time, the scalar prediction is obtained as the expectation of the bin
centres under the predicted distribution.

During training, the regression loss is automatically rescaled by the ratio
of the two base losses to put classification and regression on a similar scale without manual tuning. We
additionally use a z-loss on the query logits to prevent logit explosion and a
cross-entropy loss on context rows. The context prediction loss encourages the
representations used for context and query rows to remain similar.

\subsection{Inference Without Retrieval}

The row-based transformer, long-context training, context-dependent attention
scaling, and gated attention allow the model to use large contexts. We ran the
model without retrieval on datasets with up to 100k context rows, using a
single shared sequence containing all context rows followed by all query rows.
This removes the need to build a kNN index and avoids passing a separate
context for every query point through the model. This makes inference
substantially more efficient than TabDPT v1.1.

\section{Experiments}

We evaluate \modelname on TabArena-Lite, CC18, and CTR23. Since our goal is to provide efficient inference, we report the default setting without using the tuned or ensembled configurations reported by TabArena. Our evaluation employs 8 forward passes per dataset, matching the evaluation protocol used for top performing foundation models such as TabICLv2, the TabPFN family, and previous TabDPT versions.

In~\Cref{fig:tabarena-elo} of the Appendix, we see that, on TabArena-Lite, \modelname ranks fourth, trailing only the most recent foundation
models while outperforming TabDPT v1.1, XGBoost, TabICL, and TabPFNv2. While \modelname does not claim the top spot in raw prediction accuracy, it offers near-peak
performance and improves substantially over TabDPT v1.1 at a low inference cost.
\modelname's performance is comparable to models that are computationally costlier or need dataset-specific training.

\begin{figure}[h]
    \centering
    \includegraphics[width=0.48\textwidth]{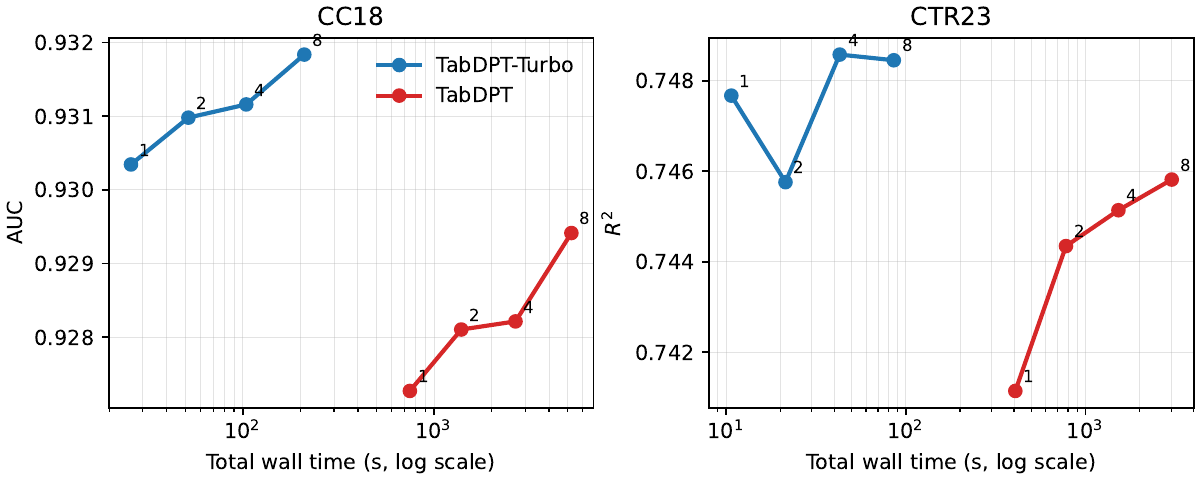}
    \caption{Performance versus total wall time for \modelname and TabDPT v1.1 on CC18 and CTR23 with $1,2,4$ and $8$ forward passes. $K=2{,}048$ neighours were used for TabDPT v1.1.}
    \label{fig:main-walltime}
    \vspace{-1em}
\end{figure}

We further compare \modelname against TabDPT v1.1 on CC18 and CTR23 in~\Cref{fig:main-walltime}. For
each model, we vary the number of inference passes and plot predictive
performance against measured wall-clock inference time. Because previous versions of TabDPT
construct a query-specific context, their inference cost grows
with the number of query points and retrieved examples. In contrast, \modelname
uses a single shared context followed by all query rows, allowing the query set
to be evaluated in one efficient, batched forward pass.

\subsection{Inference Speed}

Even with 8 forward passes per dataset, our total time for fitting and predicting is an average of 0.76s per 1000 instances. The existing TabArena-Lite results only report \(k\)-NN, ExtraTrees, and Random Forest methods as being faster, and notably report substantially slower total times for \textbf{all} gradient boosted tree ensembles or neural network models. While our hardware configurations differed, all models with comparable performance either require supervised training or use computationally heavier architectures, making it plausible for our method to be faster.

To evaluate our speed compared to other TFMs on the same hardware, we reran TabArena-Lite for \modelname as well as TabPFN-3 and TabICLv2, the fastest leading TFMs to the best of our knowledge. Results are shown in \Cref{tab:speed}, with \modelname shown to be the fastest model.

\begin{table}[th]
\centering
\caption{Average speeds on TabArena-Lite. All results generated with the same compute, using 1x H100 GPU and 96 vCPUs. All models use the default ensemble size of 8.}
\label{tab:speed}
\begin{tabular}{lrr}
\toprule
Model & \thead{Fit time \\ per 1k rows} & \thead{Predict time \\ per 1k rows} \\
\midrule
\modelname & 0.44s & 0.19s \\
TabPFN-3 & 0.87s & 0.64s \\
TabICLv2 & 0.41s & 7.75s \\
\bottomrule
\end{tabular}
\end{table}

\subsection{Ablations and Scaling}

\begin{figure}[h]
    \centering
    \includegraphics[width=0.45\textwidth]{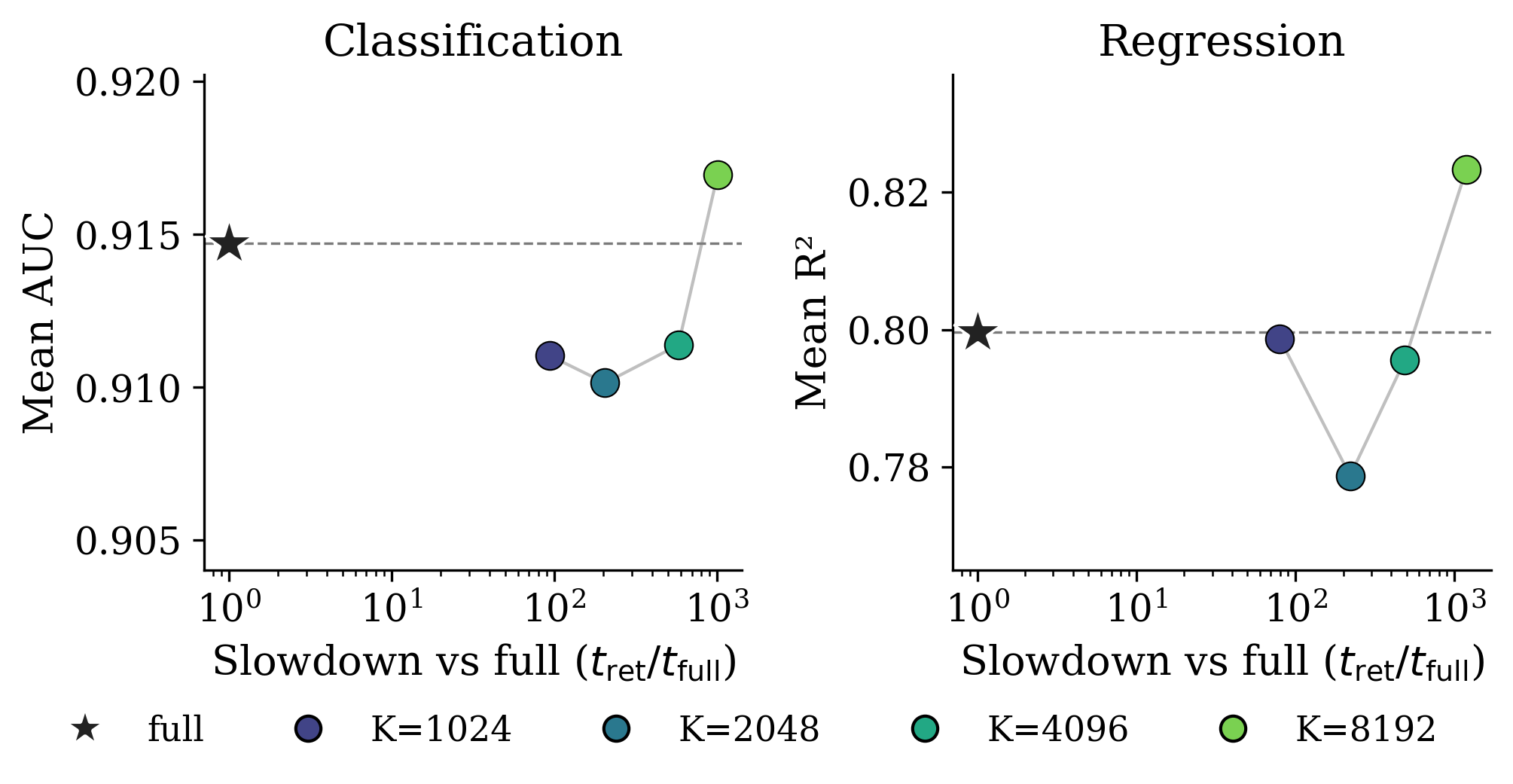}
    \caption{\looseness=-1 \textbf{Inference time vs.\ accuracy.} \modelname with no retrieval (star, dashed baseline) compared against retrieval with varying numbers of retrieved rows $K$ (circles). The x-axis shows inference slowdown relative to full-context mode ($t_{\text{ret}}/t_{\text{full}}$, log scale). Left: Mean AUC on CC18 and TabArena-Lite classification tasks. Right: Mean $R^2$ on CTR23 and TabArena-Lite regression tasks.}
    \label{fig:retrieval-tradeoff}
    \vspace{-1em}
\end{figure}

\pgraph{Retrieval Ablation}
\Cref{fig:retrieval-tradeoff} compares retrieval at varying sample sizes $K$ against full-context inference on \modelname. On both classification and regression tasks, full context outperforms retrieval at every $K$ except $K=8{,}192$, and even there the margin is small (mean AUC of 0.917 vs.\ 0.915, mean $R^2$ of 0.83 vs.\ 0.80). Meanwhile, retrieval is at least 100$\times$ slower than full-context inference, and the only configuration that surpasses full context, $K=8{,}192$, is nearly 1000$\times$ slower. This suggests that the performance gap between full-context attention and nearest-neighbour retrieval is negligible, and that retrieval no longer justifies the inference-time cost it introduces.

\pgraph{Scaling} In~\Cref{sec:app-scaling} of the Appendix, we investigate scaling both the model and context size, demonstrating that scaling both context length and parameter count leads to significant performance improvements.

\section{Conclusion}

While modern tabular ICL models have focused on maximizing predictive performance, this work pushes the boundaries of their efficiency. We present a row-based, retrieval-free, long-context model that enables computational trade-offs previously unavailable to neural network models.

Looking ahead, we plan to continue to improve the TabDPT model family in terms of predictive performance and speed, with this version providing a more effective starting point for scaling data and compute. We are also interested in leveraging it for understudied tabular data regimes, including datasets with large feature counts, large instance counts, and highly imbalanced data.

\bibliography{refs}

\begin{thebibliography}{25}
\providecommand{\natexlab}[1]{#1}
\providecommand{\url}[1]{\texttt{#1}}
\expandafter\ifx\csname urlstyle\endcsname\relax
  \providecommand{\doi}[1]{doi: #1}\else
  \providecommand{\doi}{doi: \begingroup \urlstyle{rm}\Url}\fi

\bibitem[Balazadeh~Meresht et~al.(2025)Balazadeh~Meresht, Kamkari, Thomas, Ma, Li, Cresswell, and Krishnan]{balazadeh2025causalpfn}
Balazadeh~Meresht, V., Kamkari, H., Thomas, V., Ma, J., Li, B., Cresswell, J., and Krishnan, R.
\newblock {CausalPFN}: Amortized causal effect estimation via in-context learning.
\newblock In \emph{Advances in Neural Information Processing Systems}, 2025.

\bibitem[Bischl et~al.(2021)Bischl, Casalicchio, Feurer, Gijsbers, Hutter, Lang, Gomes~Mantovani, van Rijn, and Vanschoren]{bischl2021cc18}
Bischl, B., Casalicchio, G., Feurer, M., Gijsbers, P., Hutter, F., Lang, M., Gomes~Mantovani, R., van Rijn, J., and Vanschoren, J.
\newblock {OpenML} benchmarking suites.
\newblock In \emph{Proceedings of the Neural Information Processing Systems Track on Datasets and Benchmarks}, volume~1, 2021.

\bibitem[Bischl et~al.(2025)Bischl, Casalicchio, Das, Feurer, Fischer, Gijsbers, Mukherjee, Müller, Németh, Oala, Purucker, Ravi, van Rijn, Singh, Vanschoren, van~der Velde, and Wever]{OpenML2025}
Bischl, B., Casalicchio, G., Das, T., Feurer, M., Fischer, S., Gijsbers, P., Mukherjee, S., Müller, A.~C., Németh, L., Oala, L., Purucker, L., Ravi, S., van Rijn, J.~N., Singh, P., Vanschoren, J., van~der Velde, J., and Wever, M.
\newblock {OpenML}: Insights from 10 years and more than a thousand papers.
\newblock \emph{Patterns}, 6\penalty0 (7), 2025.

\bibitem[Bouadi et~al.(2025)Bouadi, Seth, Tanna, and Sankarapu]{bouadi25orionmsp}
Bouadi, M., Seth, P., Tanna, A., and Sankarapu, V.~K.
\newblock {Orion-MSP}: Multi-scale sparse attention for tabular in-context learning.
\newblock \emph{arXiv:2511.02818}, 2025.

\bibitem[Breiman(2001)]{breiman2001random}
Breiman, L.
\newblock Random forests.
\newblock \emph{Machine learning}, 45:\penalty0 5--32, 2001.

\bibitem[Erickson et~al.(2025)Erickson, Purucker, Tschalzev, Holzm{\"u}ller, Desai, Salinas, and Hutter]{erickson2025tabarena}
Erickson, N., Purucker, L., Tschalzev, A., Holzm{\"u}ller, D., Desai, P.~M., Salinas, D., and Hutter, F.
\newblock {TabArena}: A living benchmark for machine learning on tabular data.
\newblock In \emph{Advances in Neural Information Processing Systems}, 2025.

\bibitem[Fischer et~al.(2023)Fischer, Feurer, and Bischl]{fischer2023ctr}
Fischer, S.~F., Feurer, M., and Bischl, B.
\newblock {OpenML-CTR23} -- {A} curated tabular regression benchmarking suite.
\newblock In \emph{AutoML Conference (Workshop)}, 2023.

\bibitem[Gorishniy et~al.(2025)Gorishniy, Kotelnikov, and Babenko]{gorishniy2024tabm}
Gorishniy, Y., Kotelnikov, A., and Babenko, A.
\newblock {TabM: Advancing tabular deep learning with parameter-efficient ensembling}.
\newblock In \emph{International Conference on Learning Representations}, 2025.

\bibitem[Grinsztajn et~al.(2025)Grinsztajn, Fl{\"o}ge, Key, Birkel, Jund, Roof, J{\"a}ger, Safaric, Alessi, Hayler, Manium, Yu, Jablonski, Hoo, Garg, Robertson, B{\"u}hler, Moroshan, Purucker, Cornu, Wehrhahn, Bonetto, Sch{\"o}lkopf, Gambhir, Hollmann, and Hutter]{grinsztajn2025tabpfn}
Grinsztajn, L., Fl{\"o}ge, K., Key, O., Birkel, F., Jund, P., Roof, B., J{\"a}ger, B., Safaric, D., Alessi, S., Hayler, A., Manium, M., Yu, R., Jablonski, F., Hoo, S.~B., Garg, A., Robertson, J., B{\"u}hler, M., Moroshan, V., Purucker, L., Cornu, C., Wehrhahn, L.~C., Bonetto, A., Sch{\"o}lkopf, B., Gambhir, S., Hollmann, N., and Hutter, F.
\newblock {TabPFN-2.5}: Advancing the state of the art in tabular foundation models.
\newblock \emph{arXiv: 2511.08667}, 2025.

\bibitem[Hollmann et~al.(2023)Hollmann, M{\"u}ller, Eggensperger, and Hutter]{hollmann2023tabpfn}
Hollmann, N., M{\"u}ller, S., Eggensperger, K., and Hutter, F.
\newblock {TabPFN}: A transformer that solves small tabular classification problems in a second.
\newblock In \emph{International Conference on Learning Representations}, 2023.

\bibitem[Hollmann et~al.(2025)Hollmann, Müller, Purucker, Krishnakumar, Körfer, Hoo, Schirrmeister, and Hutter]{hollmann2025accurate}
Hollmann, N., Müller, S., Purucker, L., Krishnakumar, A., Körfer, M., Hoo, S.~B., Schirrmeister, R.~T., and Hutter, F.
\newblock Accurate predictions on small data with a tabular foundation model.
\newblock \emph{Nature}, 637\penalty0 (8045):\penalty0 319--326, 2025.

\bibitem[Landsgesell \& Knoll(2026)Landsgesell and Knoll]{landsgesell2026scoringbench}
Landsgesell, J. and Knoll, P.
\newblock {ScoringBench}: A benchmark for evaluating tabular foundation models with proper scoring rules.
\newblock \emph{arXiv:2603.29928}, 2026.

\bibitem[Lou et~al.(2013)Lou, Caruana, Gehrke, and Hooker]{lou2013accurate}
Lou, Y., Caruana, R., Gehrke, J., and Hooker, G.
\newblock Accurate intelligible models with pairwise interactions.
\newblock In \emph{Proceedings of the 19th ACM SIGKDD international conference on Knowledge discovery and data mining}, pp.\  623--631, 2013.

\bibitem[Ma et~al.(2024)Ma, Thomas, Yu, and Caterini]{ma2024context}
Ma, J., Thomas, V., Yu, G., and Caterini, A.~L.
\newblock In-context data distillation with {TabPFN}.
\newblock In \emph{ICLR Workshop on Understanding of Foundation Models (ME-FoMo)}, 2024.

\bibitem[Ma et~al.(2025)Ma, Thomas, Hosseinzadeh, Kamkari, Labach, Cresswell, Golestan, Yu, Caterini, and Volkovs]{ma2025tabdpt}
Ma, J., Thomas, V., Hosseinzadeh, R., Kamkari, H., Labach, A., Cresswell, J.~C., Golestan, K., Yu, G., Caterini, A.~L., and Volkovs, M.
\newblock {TabDPT}: Scaling tabular foundation models on real data.
\newblock In \emph{Advances in Neural Information Processing Systems}, 2025.

\bibitem[Prokhorenkova et~al.(2018)Prokhorenkova, Gusev, Vorobev, Dorogush, and Gulin]{prokhorenkova2018catboost}
Prokhorenkova, L., Gusev, G., Vorobev, A., Dorogush, A.~V., and Gulin, A.
\newblock {CatBoost}: Unbiased boosting with categorical features.
\newblock In \emph{Advances in Neural Information Processing Systems}, 2018.

\bibitem[Qu et~al.(2025)Qu, Holzm{\"u}ller, Varoquaux, and Morvan]{qu2025tabicl}
Qu, J., Holzm{\"u}ller, D., Varoquaux, G., and Morvan, M.~L.
\newblock {TabICL}: A tabular foundation model for in-context learning on large data.
\newblock In \emph{International Conference on Machine Learning}, 2025.

\bibitem[Qu et~al.(2026)Qu, Holzm{\"u}ller, Varoquaux, and Morvan]{qu2026tabiclv2}
Qu, J., Holzm{\"u}ller, D., Varoquaux, G., and Morvan, M.~L.
\newblock {TabICLv2}: A better, faster, scalable, and open tabular foundation model.
\newblock In \emph{International Conference on Machine Learning}, 2026.

\bibitem[Shaheen et~al.(2026)Shaheen, Ma, Labach, Hutter, Thomas, and Caterini]{shaheen2026generalization}
Shaheen, N., Ma, J., Labach, A., Hutter, F., Thomas, V., and Caterini, A.~L.
\newblock Understanding the surprising generalization properties of tabular foundation models.
\newblock \emph{arXiv:2608.17957}, 2026.

\bibitem[Thomas et~al.(2024)Thomas, Ma, Hosseinzadeh, Golestan, Yu, Volkovs, and Caterini]{retrieve2024thomas}
Thomas, V., Ma, J., Hosseinzadeh, R., Golestan, K., Yu, G., Volkovs, M., and Caterini, A.~L.
\newblock Retrieval \& fine-tuning for in-context tabular models.
\newblock In \emph{Advances in Neural Information Processing Systems}, 2024.

\bibitem[van Breugel \& van~der Schaar(2024)van Breugel and van~der Schaar]{breugel2024tabular}
van Breugel, B. and van~der Schaar, M.
\newblock Why tabular foundation models should be a research priority.
\newblock In \emph{International Conference on Machine Learning}, 2024.

\bibitem[Vanschoren et~al.(2014)Vanschoren, Van~Rijn, Bischl, and Torgo]{vanschoren2014openml}
Vanschoren, J., Van~Rijn, J.~N., Bischl, B., and Torgo, L.
\newblock {OpenML}: Networked science in machine learning.
\newblock \emph{ACM SIGKDD Explorations Newsletter}, 15\penalty0 (2):\penalty0 49--60, 2014.

\bibitem[Veli{\v{c}}kovi{\'c} et~al.(2025)Veli{\v{c}}kovi{\'c}, Perivolaropoulos, Barbero, and Pascanu]{velickovic2025softmax}
Veli{\v{c}}kovi{\'c}, P., Perivolaropoulos, C., Barbero, F., and Pascanu, R.
\newblock Softmax is not enough (for sharp size generalisation).
\newblock In \emph{International Conference on Machine Learning}, 2025.

\bibitem[Zhang et~al.(2025{\natexlab{a}})Zhang, Maddix~Robinson, Yin, Erickson, Ansari, Han, Zhang, Akoglu, Faloutsos, Mahoney, Hu, Rangwala, Karypis, and Wang]{zhang2025mitra}
Zhang, X., Maddix~Robinson, D., Yin, J., Erickson, N., Ansari, A.~F., Han, B., Zhang, S., Akoglu, L., Faloutsos, C., Mahoney, M., Hu, T., Rangwala, H., Karypis, G., and Wang, Y.~B.
\newblock Mitra: Mixed synthetic priors for enhancing tabular foundation models.
\newblock In \emph{Advances in Neural Information Processing Systems}, 2025{\natexlab{a}}.

\bibitem[Zhang et~al.(2025{\natexlab{b}})Zhang, Ren, Yu, Yuan, Wang, Li, Wu, Mo, Mao, Hao, et~al.]{zhang2025limix}
Zhang, X., Ren, G., Yu, H., Yuan, H., Wang, H., Li, J., Wu, J., Mo, L., Mao, L., Hao, M., et~al.
\newblock {LimiX}: Unleashing structured-data modeling capability for generalist intelligence.
\newblock \emph{arXiv:2509.03505}, 2025{\natexlab{b}}.

\end{thebibliography}
\bibliographystyle{icml2026}

\newpage
\appendix
\onecolumn
\section{Appendix}

\subsection{Architecture Diagram}
As described in the main text, the architecture uses transformer blocks as its backbone. Input features are first zero-padded to the required size and passed through a linear encoder. Thinking rows are then appended to the sequence as soft tokens. Targets are routed to the transformer value projections via per-layer MLPs rather than being added to the embedded rows. The output layer is split into classification and regression heads.

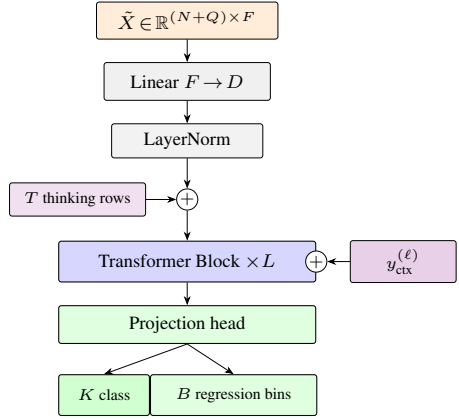
\begin{figure}[ht!]
\centering
\begin{tikzpicture}[
    font=\scriptsize,
    >={Stealth[length=1.2mm]},
    block/.style ={draw, rounded corners=1pt, minimum height=5.4mm,
                   minimum width=34mm, align=center, fill=blue!16, line width=0.4pt},
    op/.style    ={draw, rounded corners=1pt, minimum height=4mm,
                   minimum width=20mm, align=center, fill=gray!10, line width=0.3pt},
    io/.style    ={draw, rounded corners=1pt, minimum height=4.5mm,
                   minimum width=24mm, align=center, fill=orange!15, line width=0.3pt},
    head/.style  ={draw, rounded corners=1pt, minimum height=5mm,
                   minimum width=34mm, align=center, fill=green!12, line width=0.4pt},
    yctxbox/.style={draw, rounded corners=1pt, minimum height=4.4mm,
                   minimum width=14mm, align=center, fill=violet!18,
                   line width=0.3pt, font=\tiny},
    klogits/.style={draw, rounded corners=1pt, minimum height=5.4mm,
                   minimum width=12mm, align=center, fill=green!18,
                   line width=0.4pt, font=\tiny},
    blogits/.style={draw, rounded corners=1pt, minimum height=5.4mm,
                   minimum width=22mm, align=center, fill=green!12,
                   line width=0.4pt, font=\tiny},
    arr/.style   ={->, line width=0.35pt},
    sumop/.style ={draw, circle, inner sep=0.2pt, minimum size=2.8mm,
                   line width=0.3pt, fill=white, font=\tiny},
]

\node[io, anchor=north] (xin) at (0,0)
    {$\tilde X\!\in\!\mathbb{R}^{(N+Q)\times F}$};
\node[op, below=3mm of xin, minimum width=22mm, minimum height=5mm,
      fill=gray!12]                        (lin)  {Linear $F\!\to\!D$};
\node[op, below=3mm of lin, minimum width=20mm, minimum height=3.6mm]
                                            (ln)   {LayerNorm};

\node[sumop, below=4mm of ln]               (concatT) {$+$};
\node[io, fill=violet!12, left=4mm of concatT,
      minimum width=18mm, minimum height=4.4mm, font=\tiny]
                                            (think)
    {$T$ thinking rows};

\node[block, below=4mm of concatT]          (tx) {Transformer Block~$\times L$};
\node[sumop] (catY) at (tx.east) {$+$};
\node[yctxbox, right=3mm of catY]           (yctxL) {$y_{\text{ctx}}^{(\ell)}$};

\node[head, below=3mm of tx]                (projhead) {Projection head};

\node[klogits, anchor=north west]           (kpart)
    at ($(projhead.south west)+(0,-4mm)$) {$K$ class};
\node[blogits, anchor=north west]           (bpart) at (kpart.north east)
    {$B$ regression bins};

\draw[arr] (xin)     -- (lin);
\draw[arr] (lin)     -- (ln);
\draw[arr] (ln)      -- (concatT);
\draw[arr] (think)   -- (concatT);
\draw[arr] (concatT) -- (tx);
\draw[arr] (yctxL)   -- (catY);
\draw[arr] (tx)      -- (projhead);
\draw[arr] (projhead.south) -- (kpart.north);
\draw[arr] (projhead.south) -- (bpart.north);

\end{tikzpicture}
\caption{TabDPT architecture.}
\label{fig:arch-main}
\end{figure}

\subsection{Dataset comparison to TabDPT v1.1}

\begin{table}[th]
\centering
\caption{Corpus statistics: TabDPT v1.1 vs.\ TabDPT-Turbo}
\label{tab:v1_vs_v2}
\begin{tabular}{lrrr}
\toprule
Metric & TabDPT v1.1 & Ours & Ratio \\
\midrule
Number of datasets & 112     & 1{,}445  & 12.90$\times$ \\
Total rows         & 32.4M   & 309.9M   & 9.56$\times$  \\
Total features     & 15.3K   & 43.7K    & 2.86$\times$  \\
Total cells        & 0.87B   & 5.60B    & 6.41$\times$  \\
\bottomrule
\end{tabular}
\end{table}

\subsection{TabArena-Lite Elo score}

\begin{figure}[h]
    \centering
    \includegraphics[width=0.65\textwidth]{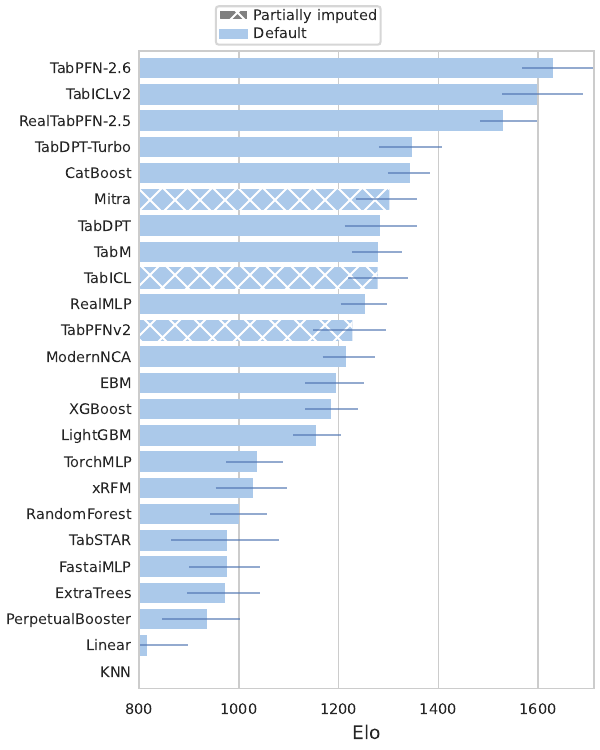}
    \caption{\textbf{TabArena-Lite Elo score comparison.} Evaluation is done with the
    default, non-tuned setting.}
    \label{fig:tabarena-elo}
\end{figure}

Figure~\ref{fig:tabarena-elo} provides the TabArena-Lite results for our models versus scores provided in the TabArena repository.

\subsection{Scaling} \label{sec:app-scaling}

We investigate scaling along two dimensions: context size and model capacity. Throughout, $d$ denotes the hidden dimension, and $L$ denotes the number of layers.

\begin{figure}[h]
    \centering
    \includegraphics[width=0.75\textwidth]{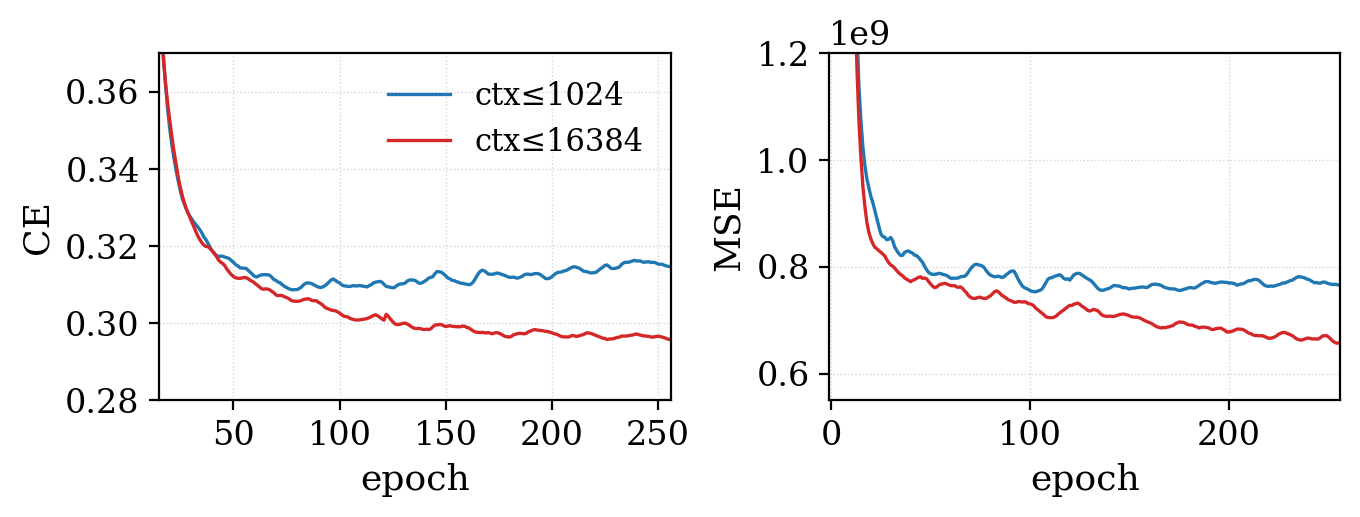}
    \caption{\textbf{Smaller vs.\ larger context.} Validation trajectories for two models with identical architecture but trained with different context sizes. Blue is trained on contexts of up to 1,024 rows and red up to 16,384.}
    \label{fig:context-scaling-curve}
    \vspace{-1em}
\end{figure}

\pgraph{Context Scaling} We trained two models of identical architecture ($d=256$, $L=8$) but with different maximum context sizes: one limited to 1,024 input rows, the other extending to 16,384. As shown in~\Cref{fig:context-scaling-curve}, the model trained with the larger context consistently outperforms its short-context counterpart, showing the benefit of increasing context size. Note that the total number of rows per batch was identical in these two settings.

\begin{figure}[h]
    \centering
    \includegraphics[width=0.75\textwidth]{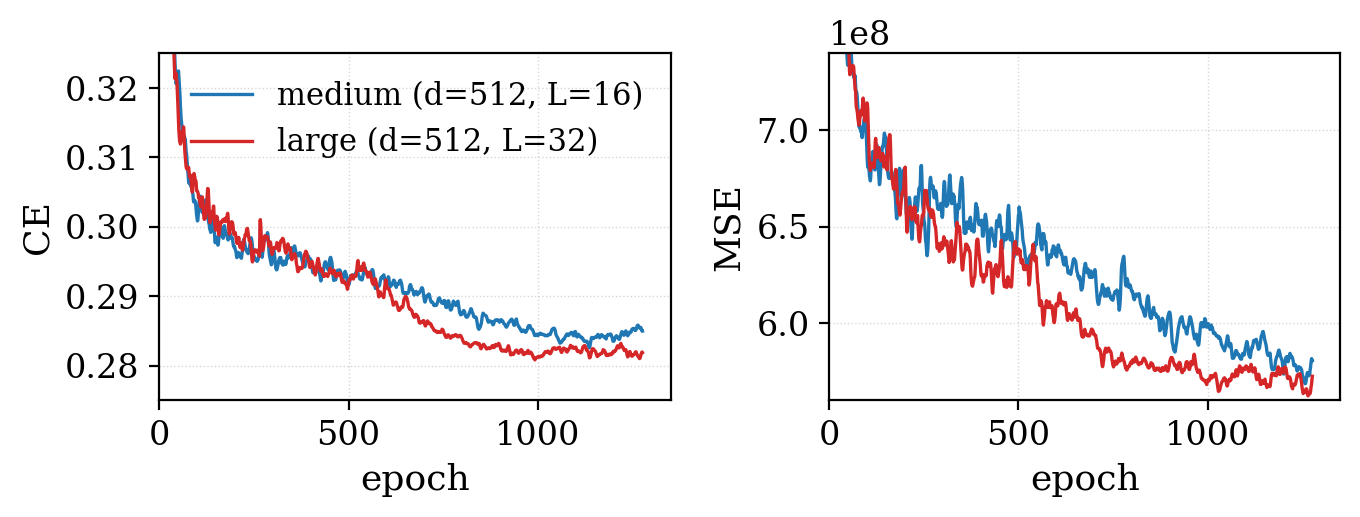}
    \caption{\textbf{Smaller vs.\ larger model.} Validation trajectories for the medium and large models. Left: cross-entropy loss on CC18 and TabArena-Lite classification tasks. Right: mean squared error on CTR23 and TabArena-Lite regression tasks.}
    \label{fig:model-scaling-curve}
\end{figure}

\pgraph{Model Scaling}
We trained a medium-sized model with half the number of attention blocks as the full \modelname, giving both models a maximum context length of 32,768 for a fair comparison. As shown in~\Cref{fig:model-scaling-curve}, \modelname (labelled ``large'') outperforms the medium model after roughly 500 epochs, indicating that scaling the parameter count yields meaningful gains.

We therefore see that scaling both context length and parameter count leads to significant performance improvements.

\end{document}